\documentclass{article}
\usepackage[preprint]{icml2026}

\usepackage{amsmath,amssymb,amsfonts}
\usepackage{booktabs}
\usepackage{graphicx}
\usepackage{hyperref}
\usepackage{url}
\usepackage{microtype}
\usepackage{xcolor}
\usepackage{multirow}
\usepackage{algorithm}
\usepackage{algorithmic}
\usepackage{subcaption}
\usepackage{float}                    %

\icmltitlerunning{From Risk Scoring to Risk Allocation in MAS}

\begin{document}

\twocolumn[
\icmltitle{From Risk Scoring to Risk Allocation: \\ A Density-Driven Framework for Diverse Monitoring in Multi-Agent Systems}

\icmlsetsymbol{equal}{*}
\begin{icmlauthorlist}
\icmlauthor{Zhaohui Wang}{usc}
\end{icmlauthorlist}
\icmlaffiliation{usc}{University of Southern California, Los Angeles, CA, USA}
\icmlcorrespondingauthor{Zhaohui Wang}{zwang000@usc.edu}
\icmlkeywords{QUBO, subset selection, multi-agent systems, risk monitoring, QAOA}

\vskip 0.3in
]

\printAffiliationsAndNotice{}

\begin{abstract}
Risk monitoring in multi-agent systems is commonly built on a per-state primitive that scores each state independently and selects the top $K$.
Under crowding, where many agents share the same fragility, this approach picks redundant alerts whose risks are jointly correlated, a pattern we describe as ``herding in monitoring.''
We propose a paradigm shift from risk scoring to risk allocation, supported by two contributions.
First, we identify the \emph{Crowding Paradox}, namely that $P(\text{risk}\mid x)\propto p(x)$ rather than $1/p(x)$, so density rather than anomaly score is the operative risk signal; on financial data, density-based scoring reaches AUROC $\ge 0.94$ at 5d/10d/20d crash horizons, while five anomaly baselines all fall below $0.80$.
Second, given a density-derived fragility score, we recast monitoring as combinatorial subset selection over interdependent states and map it to a QUBO objective with a $\lambda$-controlled risk--diversity tradeoff.
The resulting Pareto frontier contains standard diverse-subset methods (MMR, $k$-DPP) as fixed operating points; the gain over greedy grows monotonically with scale, from $+24\%$ at $n{=}15$ to $+66\%$ at $n{=}200$; a learned $\lambda$ policy reaches $99.5\%$ of an oracle grid-search objective; and the formulation transfers to traffic and multi-agent reinforcement learning.
The same QUBO instances execute without modification on Rigetti superconducting QPUs (Ankaa-3 and Cepheus-1-108Q via Amazon Braket), which we report as a compatibility property of the formulation rather than a claim of quantum advantage at this scale.
\end{abstract}

\section{Introduction}
\label{sec:intro}

A central feature of risk monitoring in multi-agent systems is its anomaly-centric primitive: states are scored independently, and the lowest-probability states are flagged for attention.
This primitive is at odds with how systemic risk is generated.
In multi-agent systems such as financial markets, autonomous fleets, and cooperative robotics, risk concentrates not in rare states but in crowded ones, where many agents adopt similar strategies and share the same failure modes.
Banerjee's herding model~\citep{banerjee1992simple} shows that rational agents may ignore private information and follow their predecessors, creating information cascades.
The 2007 quant meltdown provides direct evidence: quantitative funds with similar factor exposures experienced synchronized liquidations~\citep{khandani2011what}, and subsequent studies confirm that crowding measures predict tail risk across asset classes~\citep{brown2022crowded,kang2021crowding}, with fire-sale contagion amplifying losses when crowded positions unwind simultaneously~\citep{cont2016fire,greenwood2015vulnerable}.
These findings point to what we call the \textbf{Crowding Paradox}: in contrast to anomaly-detection methods that equate risk with low-probability states~\citep{chandola2009anomaly}, systemic risk arises from \emph{high-density} states where agents cluster, so that $P(\text{risk}\mid x) \propto p(x)$ rather than $1/p(x)$.
We validate this empirically in App.~\ref{app:density_auroc}: on the 2024--2026 financial test set, GMM density and the derived Fragility Score reach AUROC values of $0.94$, $0.94$, and $0.96$ at 5d, 10d, and 20d crash horizons respectively, while five anomaly-detection baselines (Isolation Forest, OCSVM, LOF, PCA-reconstruction, and AE-reconstruction) all score below $0.80$ AUROC and below $0.06$ AUPRC.

Identifying which states are fragile is only half of the problem; the remaining half is allocating finite monitoring capacity across them.
This motivates the central reframing of the paper, namely a shift from risk scoring to risk allocation.
Score-and-threshold approaches implicitly assume that the $K$ riskiest states are statistically independent, an assumption that fails under crowding because crowded states share failure modes.
The resulting top-$K$ selection clusters in feature space and is vulnerable to a single shock, an outcome that we describe as ``herding in monitoring'' since it mirrors the herding that creates fragility in the first place.

Subset selection with a diversity constraint is a \emph{discrete optimization} problem.
Rather than relying on ad-hoc post-processing such as diversity re-ranking, we formulate the problem directly as QUBO~\citep{glover2022tutorial}:
\begin{multline}
\min_{\mathbf{z} \in \{0,1\}^n} \; \underbrace{-\textstyle\sum_i \text{FS}(x_i)\, z_i}_{\text{risk coverage}} \;+\; \lambda \underbrace{\textstyle\sum_{i<j} S_{ij}\, z_i z_j}_{\text{diversity}} \\
+\; P\!\left(\textstyle\sum_i z_i - K\right)^{\!2}
\label{eq:qubo}
\end{multline}
where $\text{FS}(x_i)$ is the Fragility Score of state $x_i$, $S_{ij}$ measures pairwise similarity, and $P$ enforces cardinality $K$.

This formulation offers three advantages over score-and-threshold approaches:
(i)~it jointly optimizes risk and diversity in a single objective;
(ii)~it is agnostic to the solver backend, so that classical heuristics, exact methods, and quantum processors can all solve the same QUBO;
(iii)~it provides a principled $\lambda$-parameterized tradeoff between risk concentration and coverage diversity.

\begin{figure*}[!t]
\centering
\includegraphics[width=\textwidth]{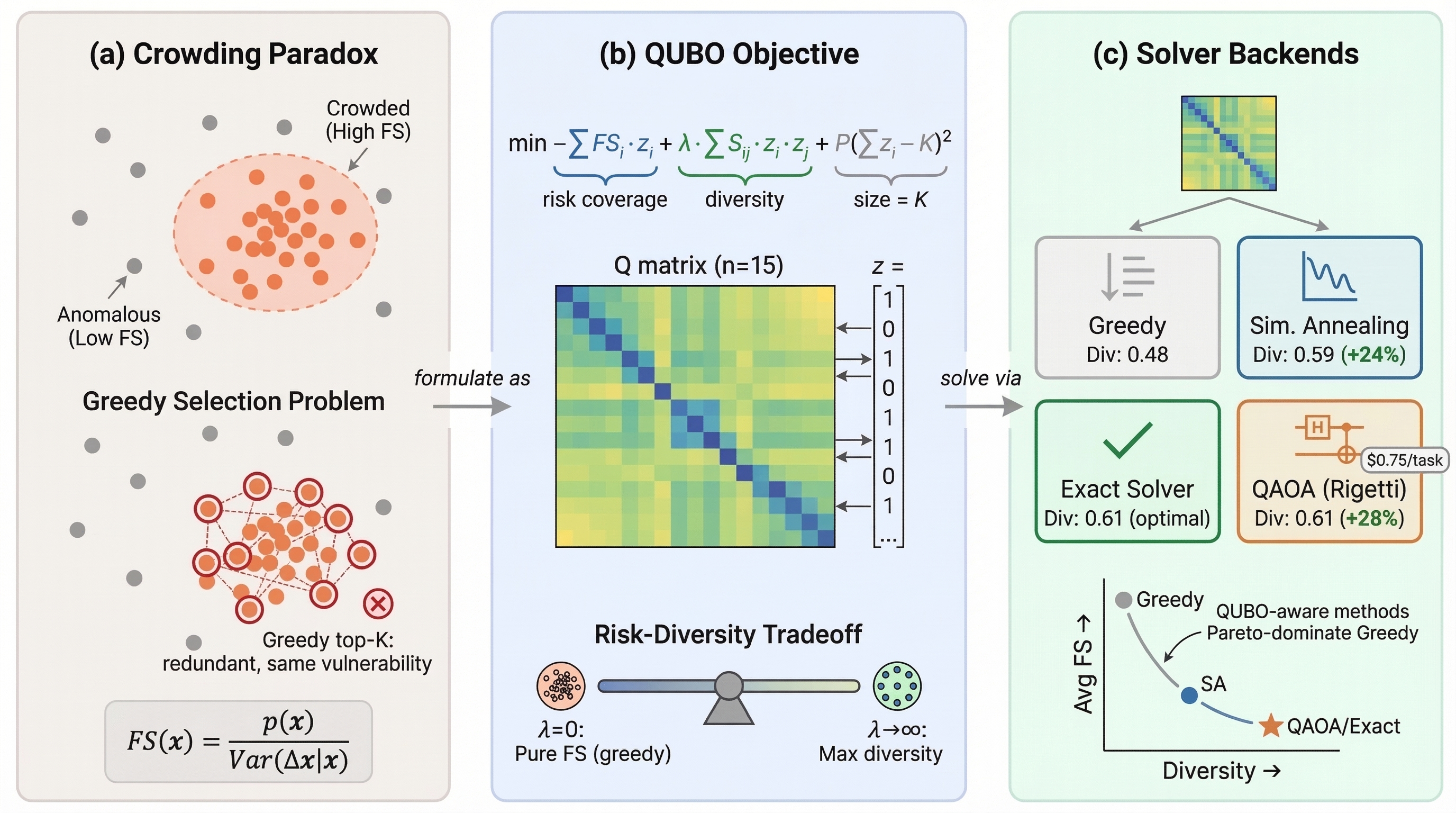}
\caption{From scoring to allocation. Unlike scoring-based pipelines that threshold each state independently, the proposed formulation treats monitoring as a \emph{joint decision} over interdependent states.
(a)~Multi-agent crowding creates high-density, low-variance states (high Fragility Score); greedy top-$K$ selection picks redundant states sharing the same vulnerability.
(b)~The QUBO objective jointly optimizes risk coverage (blue), diversity (green), and cardinality (gray) in a single quadratic form, with $\lambda$ controlling the risk-diversity tradeoff.
(c)~The same QUBO instance is solved by four backends: greedy (Div~0.48), simulated annealing (Div~0.59, $+24\%$), exact solver (optimal), and QAOA on Rigetti hardware. All QUBO-aware methods Pareto-dominate greedy.}
\label{fig:overview}
\end{figure*}

\paragraph{Contributions.}
\textbf{(1) Conceptual.} We identify and empirically validate the \emph{Crowding Paradox}, namely that in multi-agent systems risk concentrates in high-density states rather than anomalies, so that density rather than anomaly score is the operative risk signal (App.~\ref{app:density_auroc}).
\textbf{(2) Reframing.} We recast risk-aware monitoring as a combinatorial allocation problem over interdependent states, rather than as independent per-state scoring (\S\ref{sec:formulation}).
\textbf{(3) Formulation.} We show that this allocation maps to QUBO, yielding a single quadratic objective that any QUBO-compatible solver, including classical heuristics, exact methods, quantum annealers, and gate-model algorithms, can attack without reformulation (\S\ref{sec:formulation},\,\S\ref{sec:solvers}).
\textbf{(4) Empirical.} The formulation produces a controllable risk--diversity Pareto frontier on which standard methods (greedy, MMR, $k$-DPP) sit as fixed points; the gain over greedy grows with scale, from $+24\%$ at $n{=}15$ to $+66\%$ at $n{=}200$; the formulation generalizes across finance, traffic, and multi-agent reinforcement learning; and the same QUBO instances are executable on real Rigetti QPUs (\S\ref{sec:experiments}, App.~\ref{app:scaling}--\ref{app:downstream}).

The substantive contribution of this work is the reframing of risk monitoring rather than the QUBO encoding itself; the QUBO formulation is used because it is the smallest quadratic encoding that supports the reframing without auxiliary variables or continuous relaxation.

\section{Problem Formulation}
\label{sec:formulation}

\paragraph{Setting.}
Consider a multi-agent system producing a time series of market states $\{x_t\}_{t=1}^T$, where $x_t \in \mathbb{R}^d$ encodes features such as returns, volatilities, and cross-asset correlations.
A density model (GMM with $k{=}5$ components) trained on historical data provides $p(x_t)$, and conditional variance $\text{Var}(\Delta x | x_t)$ is estimated from a rolling 20-day window.
High-density states with suppressed variance are most vulnerable: agents in crowded positions reduce activity, compressing variance while fragility accumulates~\citep{brown2022crowded}.
We capture this mechanism with the \textbf{Fragility Score}:
\begin{equation}
\text{FS}(x) = \frac{p(x)}{\text{Var}(\Delta x | x)}
\label{eq:fs}
\end{equation}
States with high density but low variance, the ``calm before the storm,'' receive the highest fragility scores.
For numerical stability, we rank-normalize FS to $[0,1]$ across the full test set.

\paragraph{Candidate pool.}
From the test set ($T{=}523$ trading days), we select the top-$n$ states by FS as the candidate pool.
We use $n{=}15$ as the primary setting to enable exact enumeration as ground truth, allowing controlled comparison across solvers under identical QUBO instances and isolating solver behavior from approximation error.
We also evaluate scaling to $n{=}20$, $25$, and $30$ (\S\ref{app:scaling}).
The monitoring task is to choose $K{=}8$ states that balance high aggregate fragility with diversity.

\paragraph{QUBO construction.}
The QUBO formulation is not an arbitrary encoding but the smallest quadratic form that simultaneously expresses (i) the additive risk-coverage reward $\sum_i \text{FS}(x_i) z_i$, (ii) the pairwise diversity penalty $\sum_{i<j} S_{ij} z_i z_j$, and (iii) the exact cardinality constraint $\sum_i z_i = K$ as a polynomial, with no higher-order term, no auxiliary variable, and no continuous relaxation.
We normalize FS within the candidate pool to $[0,1]$ and compute pairwise similarity $S_{ij}$ via an RBF kernel on the $d$-dimensional feature vectors, also normalized to $[0,1]$.
The QUBO matrix $Q$ is obtained by expanding Eq.~\ref{eq:qubo}:
\begin{align}
Q_{ii} &= -\widehat{\text{FS}}_i + P(1-2K) \\
Q_{ij} &= \lambda S_{ij} + 2P \quad (i < j)
\end{align}
where $P{=}2.0$ is the cardinality penalty strength and $\lambda \in [0, 2]$ controls the risk-diversity tradeoff.
The constant $P$ is chosen so that the marginal cardinality penalty $2P=4$ exceeds $\lambda \cdot \max_{i<j} S_{ij} \le \lambda$ for $\lambda \le 2$, which guarantees that no swap that violates $|\mathcal{S}|=K$ can lower the objective at $\lambda \le 0.5$; at $\lambda{=}1.0$ the bound is tight and we observe a small relaxation of the cardinality (mean $|\mathcal{S}|=7.2$ in Table~\ref{tab:main}), indicating the soft penalty is approaching its operating limit.
When $\lambda{=}0$, the problem reduces to pure fragility maximization (equivalent to greedy); as $\lambda$ increases, the solver trades fragility for diversity.
This QUBO can equivalently be expressed as an Ising Hamiltonian via $z_i = (1+s_i)/2$~\citep{lucas2014ising}, enabling direct mapping to quantum hardware.

\paragraph{Generalization of standard methods (informal).}
Greedy top-$K$ selection corresponds to the special case $\lambda{=}0$ in Eq.~\ref{eq:qubo}, since the diversity term vanishes and the cardinality penalty enforces $|\mathcal{S}|=K$. Maximal Marginal Relevance with diversity weight $\mu$ corresponds, under iterative greedy decoding, to a $\lambda$-dependent surrogate of the same objective with effective $\lambda \approx \mu/(1-\mu)$. The $k$-DPP MAP problem under the quality-similarity kernel $L_{ij} = \widehat{\text{FS}}_i \widehat{\text{FS}}_j S_{ij}$ admits a log-determinant relaxation that, locally near the QUBO minimum, is dominated by the same pairwise similarity term. Greedy, MMR, and $k$-DPP therefore correspond to specific (sometimes implicit) $\lambda$ values of a single QUBO objective, while the QUBO admits the full continuum $\lambda \in [0,2]$, and so the formulation is a generalization of the three. Our experiments confirm that MMR$(\mu)$ and $k$-DPP land on the SA Pareto curve (Tables~\ref{tab:diverse_baselines},~\ref{tab:downstream}).

\section{Solvers}
\label{sec:solvers}

We compare four solvers on identical QUBO instances.
Algorithm~\ref{alg:pipeline} summarizes the end-to-end pipeline.

\begin{algorithm}[t]
\caption{QUBO-based fragility-aware subset selection}
\label{alg:pipeline}
\begin{algorithmic}[1]
\REQUIRE Test states $\{x_t\}$, density model $p$, variance estimates, target $K$, trade-off $\lambda$
\STATE Compute $\text{FS}(x_t)$ for all test states via Eq.~\ref{eq:fs}
\STATE Select top-$n$ by FS as candidate pool $\mathcal{C}$
\STATE Compute similarity matrix $S_{ij} = \text{RBF}(x_i, x_j)$ for $x_i, x_j \in \mathcal{C}$
\STATE Construct QUBO matrix $Q$ with parameters $\lambda, P, K$
\STATE Solve $\mathbf{z}^* = \arg\min_{\mathbf{z}} \mathbf{z}^\top Q\, \mathbf{z}$ via chosen solver
\STATE \textbf{return} Selected subset $\mathcal{S} = \{x_i \in \mathcal{C} : z_i^* = 1\}$
\end{algorithmic}
\end{algorithm}

\textbf{Greedy top-$K$} selects the $K$ highest-FS candidates, ignoring the QUBO structure entirely.
This represents current practice in risk monitoring systems.

\textbf{Simulated Annealing (SA)} uses the D-Wave Ocean SDK's \texttt{SimulatedAnnealingSampler} with 200 reads per instance~\citep{kirkpatrick1983optimization,dwave2024ocean}.

\textbf{Exact solver} enumerates all $2^n$ bitstrings for $n{\le}20$ via \texttt{dimod.ExactSolver}, providing the ground-truth optimal solution for verification.

\textbf{QAOA on quantum hardware} (App.~\ref{app:pareto},\,\ref{app:cost},\,\ref{app:analysis}). We additionally execute the same QUBO on the local Braket statevector simulator and on Rigetti superconducting QPUs (Ankaa-3, retired during this work, and its successor Cepheus-1-108Q) via Amazon Braket, using $p{=}1$ QAOA~\citep{farhi2014quantum,zhou2020qaoa} with 500 shots per submission. We report this as a compatibility check, not a quantum-advantage claim.

\section{Experiments}
\label{sec:experiments}

\paragraph{Data.}
We use daily financial market data (2018--2025) comprising 18 features: equity returns and volume changes (SPY, QQQ), Treasury yields (2Y, 3M, 10Y), yield curve slope, investment-grade and high-yield credit spreads, realized volatility (5/10/20-day windows), gold and bond returns (TLT, GLD), and cryptocurrency returns (BTC, ETH).
A GMM ($k{=}5$, full covariance) is trained on data through 2021; the test set comprises 523 trading days after 2023-12-31, with no temporal leakage.
The top 15 states by rank-normalized Fragility Score form the candidate pool.

\paragraph{Evaluation metrics.}
For each selected subset $\mathcal{S}$:
(i)~\textbf{Avg FS}: mean Fragility Score (higher $=$ more risk-aware);
(ii)~\textbf{Diversity}: $1 - \bar{S}$, where $\bar{S}$ is the mean pairwise RBF similarity within $\mathcal{S}$ (higher $=$ more diverse);
(iii)~\textbf{$|\mathcal{S}|$}: number of selected states (target: $K{=}8$).
We sweep $\lambda \in \{0.0, 0.5, 1.0\}$ and report results across 5 random seeds (80\% candidate overlap) for stability.
Detailed QAOA protocol (5 classical-optimizer seeds for the simulator; 3 independent QPU submissions per $\lambda$ on Cepheus-1-108Q; Ankaa-3 single-seed retained for the $n{=}20$ scaling check) is in App.~\ref{app:pareto}.

\subsection{Main Results}

\begin{table}[t]
\caption{Classical solver comparison at $n{=}15$, $K{=}8$, horizon${=}$5d (mean$\pm$std over 5 data-resampling seeds; Greedy is $\lambda$-independent). QAOA on simulator and on Rigetti hardware (Ankaa-3, Cepheus-1-108Q) tracks the same Pareto frontier within hardware noise; full QAOA results in Table~\ref{tab:pareto_full} (App.~\ref{app:pareto}).}
\label{tab:main}
\vskip 0.1in
\centering
\small
\begin{tabular}{@{}llccc@{}}
\toprule
$\lambda$ & Method & Avg FS & Diversity & $|\mathcal{S}|$ \\
\midrule
\multirow{3}{*}{0.0}
& Greedy & .367\scriptsize$\pm$.002 & .476\scriptsize$\pm$.035 & 8.0 \\
& SA     & .366\scriptsize$\pm$.002 & .487\scriptsize$\pm$.029 & 8.0 \\
& Exact  & .367\scriptsize$\pm$.002 & .476\scriptsize$\pm$.035 & 8.0 \\
\midrule
\multirow{3}{*}{0.5}
& Greedy & .367\scriptsize$\pm$.002 & .476\scriptsize$\pm$.035 & 8.0 \\
& SA     & .348\scriptsize$\pm$.025 & \textbf{.591}\scriptsize$\pm$.070 & 8.0 \\
& Exact  & .348\scriptsize$\pm$.025 & \textbf{.591}\scriptsize$\pm$.070 & 8.0 \\
\midrule
\multirow{3}{*}{1.0}
& Greedy & .367\scriptsize$\pm$.002 & .476\scriptsize$\pm$.035 & 8.0 \\
& SA     & .337\scriptsize$\pm$.032 & \textbf{.639}\scriptsize$\pm$.062 & 7.2 \\
& Exact  & .337\scriptsize$\pm$.032 & \textbf{.639}\scriptsize$\pm$.062 & 7.2 \\
\bottomrule
\end{tabular}
\vskip -0.1in
\end{table}

Table~\ref{tab:main} presents the core results.

\textbf{QUBO-aware methods improve diversity over both greedy selection and standard diverse-subset baselines.}
At $\lambda{=}0.5$, SA and Exact increase diversity from $0.476$ to $0.591$, a relative gain of $+24\%$ over Greedy at a cost of only $5\%$ in FS. At $\lambda{=}1.0$, diversity reaches $0.639$, a gain of $+34\%$.
Beyond Greedy, we compare against standard diverse-subset methods in Table~\ref{tab:diverse_baselines}, namely Maximal Marginal Relevance (MMR) at three diversity weights, $k$-DPP greedy MAP under the quality-similarity kernel $L_{ij}=\widehat{\text{FS}}_i \widehat{\text{FS}}_j S_{ij}$, and a score-then-rerank baseline that takes the top-$2K$ by FS and applies MMR.
The strongest of these, MMR at diversity weight $0.7$, reaches diversity $0.562$, which SA exceeds by $+5\%$ at $\lambda{=}0.5$ and by $+14\%$ at $\lambda{=}1.0$ at comparable Avg FS, with $+19\%$ over $k$-DPP.
The remaining gain over MMR and $k$-DPP is smaller than the gain over Greedy, but constitutes a Pareto improvement on the evaluated instances and reduces the chance that selected states share an underlying shock.

\begin{table}[t]
\caption{Diverse-subset baselines at $n{=}15$, $K{=}8$, horizon${=}$5d (5 seeds; mean$\pm$std). MMR$(\mu)$ uses diversity weight $\mu \in [0,1]$.}
\label{tab:diverse_baselines}
\vskip 0.1in
\centering
\small
\begin{tabular}{@{}lccc@{}}
\toprule
Method & Avg FS & Diversity & $|\mathcal{S}|$ \\
\midrule
MMR$(0.3)$        & .365\scriptsize$\pm$.002 & .510\scriptsize$\pm$.052 & 8.0 \\
MMR$(0.5)$        & .364\scriptsize$\pm$.002 & .518\scriptsize$\pm$.043 & 8.0 \\
MMR$(0.7)$        & .349\scriptsize$\pm$.024 & .562\scriptsize$\pm$.067 & 8.0 \\
$k$-DPP (greedy MAP) & .360\scriptsize$\pm$.008 & .539\scriptsize$\pm$.057 & 8.0 \\
Score+MMR$(0.5)$  & .364\scriptsize$\pm$.002 & .518\scriptsize$\pm$.043 & 8.0 \\
\midrule
\textit{QUBO SA, $\lambda{=}0.5$} & .348\scriptsize$\pm$.025 & .591\scriptsize$\pm$.070 & 8.0 \\
\textit{QUBO SA, $\lambda{=}1.0$} & .337\scriptsize$\pm$.032 & .639\scriptsize$\pm$.062 & 7.2 \\
\bottomrule
\end{tabular}
\vskip -0.1in
\end{table}

\textbf{The gain grows with problem scale.}
Extending the candidate pool to the full financial test set, SA at $\lambda{=}0.5$ achieves diversity $0.851$ at $n{=}200$ versus Greedy's $0.514$, an improvement of $+66\%$ (Table~\ref{tab:scaling_main}; details in App.~\ref{app:scaling}).
The relative gain rises monotonically with $n$, indicating that the advantage is a scale-aware property of the formulation rather than a small-$n$ artifact, and SA runtime remains practical at $56$ seconds for $n{=}200$.

\begin{table}[t]
\caption{Scaling of the SA-over-Greedy diversity gain on the full financial test set ($\lambda{=}0.5$, $K{=}8$, horizon${=}$5d). The relative gain grows monotonically with $n$; details in App.~\ref{app:scaling}.}
\label{tab:scaling_main}
\vskip 0.05in
\centering
\small
\begin{tabular}{@{}lcccc@{}}
\toprule
$n$ & 15 & 50 & 100 & 200 \\
\midrule
Greedy Div                & .476 & .495 & .505 & .514 \\
SA Div ($\lambda{=}0.5$)  & .591 & .649 & .738 & \textbf{.851} \\
Gain over Greedy          & $+24\%$ & $+31\%$ & $+46\%$ & $\mathbf{+66\%}$ \\
SA runtime                & 44ms & 0.6s & 4.1s & 56s \\
\bottomrule
\end{tabular}
\vskip -0.1in
\end{table}

\textbf{SA recovers the exact optimum, and standard methods occupy fixed points on the resulting frontier.}
At $n{=}15$, SA produces solutions identical to those of the exact solver within seed variance, so the QUBO objective defines the Pareto frontier rather than approximating it heuristically.
MMR and $k$-DPP sit on the same frontier as fixed operating points (Table~\ref{tab:diverse_baselines}): Greedy corresponds to $\lambda{=}0$, MMR$(0.7)$ to approximately $\lambda{\approx}0.3$, and $k$-DPP to approximately $\lambda{\approx}0.4$.
The QUBO formulation thus produces a controllable continuum that contains these existing methods rather than a single isolated operating point.

\textbf{The same QUBO instance executes on real quantum hardware without modification.}
We additionally verify that the QUBO instances accepted by SA are accepted, unchanged, by gate-model quantum hardware: multi-seed QAOA on Rigetti Cepheus-1-108Q yields diversity $0.592\pm 0.020$ at $\lambda{=}0.5$ and $0.616\pm 0.007$ at $\lambda{=}1.0$, both tracking the classical Pareto frontier within hardware noise.
We do not claim quantum advantage at this scale; we claim only computational compatibility, which is a property of the QUBO abstraction (details and cost itemization in App.~\ref{app:pareto},~\ref{app:cost},~\ref{app:analysis}).

\subsection{Robustness, Generalization, and Downstream Capture}
\label{sec:robustness_summary}

The single-table headline above is supported by an extended set of analyses, all reported in detail in the appendix:

\begin{itemize}\itemsep -2pt
\item \textbf{Risk--diversity Pareto front} (App.~\ref{app:pareto}): QUBO-aware methods Pareto-dominate Greedy across $\lambda \in \{0,0.5,1\}$; QAOA on both Rigetti devices tracks the SA/Exact frontier within hardware noise.
\item \textbf{Cross-horizon consistency} (App.~\ref{app:horizons}): SA/Exact solutions are identical across 5d/10d/20d horizons (the QUBO objective is horizon-agnostic).
\item \textbf{Scaling} (App.~\ref{app:scaling}): from $n{=}15$ to $n{=}200$, SA runtime grows 44ms~$\to$~56s and the diversity gain over Greedy grows monotonically from $+24\%$ to $+66\%$.
\item \textbf{Sensitivity to $K$ and $\lambda$} (App.~\ref{app:ksens}): the SA-over-Greedy advantage holds across $K\in\{4,\dots,12\}$; a fine-grained $\lambda$ sweep shows a sharp diversity onset near $\lambda\!\approx\!0.1$ and saturation beyond $\lambda\!\approx\!1$.
\item \textbf{Learned $\lambda$ policy} (App.~\ref{app:learn}): an MLP policy mapping pool statistics $\to \lambda$ achieves $99.5\%$ of the oracle grid-search objective on held-out test data.
\item \textbf{Cross-domain transfer} (App.~\ref{app:cross}): on inD traffic and SMAC v2 MARL, SA recovers diversity that Greedy fails to deliver (Greedy diversity $\to 0$ on traffic; +7\% on MARL where Greedy is already diverse).
\item \textbf{Downstream stress capture} (App.~\ref{app:downstream}): on forward-looking SPY drawdown, realized volatility, temporal coverage, and labeled structural-break recall, SA at $\lambda{=}0.5$ more than doubles structural-break recall over Greedy ($.13$ vs.\ $.05$), while MMR$(0.7)$ leads on realized drawdown ($-1.22\%$ vs.\ $-0.87\%$), an observation we interpret as complementary regimes within the same Pareto frontier.
\item \textbf{Solver computational cost} (App.~\ref{app:cost}): SA \$0 / 45ms; QAOA-Rig 1 task = \$0.75 (Ankaa-3) or \$0.51 (Cepheus-1) at 500 shots.
\end{itemize}

The remaining narrative claims of this paper -- that the formulation is solver-agnostic, the diversity gain is not an artifact of small $n$ or specific $K$, the policy generalizes across regimes, the formulation transfers across domains, and the gain over Greedy is not a tautological self-comparison -- each rest on one of these appendix sections.

\section{Related Work}
\label{sec:related}

\paragraph{Risk in multi-agent systems.}
Herding and crowding as systemic risk drivers are well established in behavioral economics~\citep{banerjee1992simple,khandani2011what,brown2022crowded,kang2021crowding}.
The density-risk relationship has been documented empirically in financial markets~\citep{brown2022crowded,kang2021crowding}, theoretically through herding models~\citep{banerjee1992simple}, and via fire-sale contagion mechanisms~\citep{cont2017fire_sale}.
We extend this line of work from prediction to decision.

\paragraph{QUBO and combinatorial optimization.}
QUBO provides a unified framework for NP-hard combinatorial problems~\citep{glover2022tutorial,lucas2014ising}, with the underlying Ising formulation known to be NP-hard~\citep{barahona1982computational}.
Applications include feature selection~\citep{mucke2023feature}, portfolio optimization~\citep{mugel2022dynamic,morapakula2025endtoend}, and community detection~\citep{negre2020detecting}, with recent work on QUBO encoding strategies~\citep{desantis2026optimized}.
Portfolio QUBO work optimizes returns, whereas we optimize monitoring diversity under a fragility-based risk model, a complementary decision problem that extends beyond finance to general MAS.

\paragraph{Quantum optimization.}
Quantum annealing~\citep{kadowaki1998quantum} and QAOA~\citep{farhi2014quantum,zhou2020qaoa} provide QUBO-native solvers in the NISQ era~\citep{preskill2018quantum}.
Recent benchmarks show D-Wave hybrid solvers competitive with classical methods on dense instances~\citep{kim2025quantum}, and quantum computing for finance is an active application area~\citep{herman2023quantum,abbas2024quantum}.
We contribute a concrete MAS application validated on real quantum hardware (Rigetti Ankaa-3 and Cepheus-1-108Q), with classical baselines for direct comparison and multi-seed hardware error bars.

\section{Limitations and Future Work}
\label{sec:limitations}

\textbf{Problem scale.}
Our controlled experiments use $n{=}15$ to enable exact verification, with scaling studies up to $n{=}30$ (SA) and $n{=}20$ (Rigetti QPU).
The practical relevance of QUBO-based decision making and quantum backends emerges at larger scales ($n{>}50$), where exact enumeration is infeasible and classical heuristics may degrade.
Quantum and hybrid approaches may provide advantages in such regimes.

\textbf{QAOA depth.}
We use $p{=}1$ (single QAOA layer), which limits expressivity.
Higher $p$ improves solution quality in theory~\citep{zhou2020qaoa} but increases circuit depth and noise sensitivity on current hardware.

\textbf{Static candidate pool.}
The current formulation selects from a fixed pool.
In practice, MAS monitoring requires rolling temporal windows in which the candidate pool and similarity structure evolve.
Extending QUBO to warm-started sequential optimization is an open direction.

\textbf{Cross-domain validation depth.}
Our deepest empirical analysis (lambda/K sweeps, scaling, learned policy, QPU runs) is on financial markets, where labeled crash horizons are available.
The cross-domain results in Sec.~\ref{sec:experiments} on inD traffic and SMAC v2 MARL are smaller-scale demonstrations of formulation transfer, not full-protocol replications; in particular, MARL is evaluated on 70 episodes.
Operational-grade validation in those domains, with domain-specific labels, larger episode pools, and rolling candidate updates, is left to future work.

\section{Conclusion}
\label{sec:conclusion}

We have shown that fragility-aware monitoring in multi-agent systems can be formulated as a QUBO problem, providing a principled, solver-agnostic framework for discrete risk allocation.
The formulation jointly optimizes risk coverage and diversity in a single quadratic objective without continuous relaxation.
We validated it with four solvers, including multi-seed QAOA on Rigetti superconducting hardware (Ankaa-3 single-seed and Cepheus-1-108Q multi-seed; \$9.83 total), all producing competitive solutions.
The $\lambda$-parameterized risk-diversity tradeoff enables operators to balance monitoring breadth against risk concentration explicitly.
These results establish QUBO as a natural computational abstraction for this class of MAS decision problems, with a clear path to quantum acceleration as hardware scales.

\section*{Impact Statement}

This work proposes a computational framework for risk monitoring in multi-agent systems.
The primary intended application is systemic risk surveillance in financial markets, where improved diversity in monitoring coverage could help detect a broader range of failure modes.
We do not foresee direct negative societal consequences from this methodology.
The quantum computing component is validated at small scale and positioned as a forward-looking capability, not a recommendation for immediate deployment.
All experiments use publicly available market data and cloud-accessible quantum hardware, ensuring reproducibility.

\section*{Reproducibility}

Code, data pipelines, and experiment scripts are in the supplementary material; classical solvers (Greedy, SA, Exact) need only \texttt{dwave-ocean-sdk} on any CPU, QAOA simulation needs \texttt{amazon-braket-sdk} (free), and the QPU runs need an AWS Braket account. Total QPU cost: \$9.83 (18 tasks); recommended device for re-running is Rigetti Cepheus-1-108Q (Ankaa-3 was retired during this work).

\bibliography{references}
\bibliographystyle{icml2026}

\appendix
\onecolumn
\section{Density vs.\ Anomaly Baselines for Crash Prediction}
\label{app:density_auroc}

This appendix supports the Crowding Paradox claim in Section~\ref{sec:intro}: that systemic-risk-relevant states are \emph{high-density} states, not low-density anomalies.
On the same 2024-01-02 to 2026-02-02 financial test set used throughout the paper (523 trading days; train: 2017-2021), we compare GMM density and the derived Fragility Score against five standard anomaly-detection baselines.
Crash labels are forward-looking $H$-day max drawdown beyond a $-3\sigma$ threshold; positive counts are 4 ($H{=}5$), 9 ($H{=}10$), and 24 ($H{=}20$).

\begin{table}[H]
\caption{Crash-prediction AUROC and AUPRC for density-based scoring vs.\ anomaly-detection baselines on the financial test set. Density-based methods (top two rows) treat \emph{high} probability as risk-relevant; anomaly methods (lower five) treat \emph{low} probability as risk-relevant. The reversal in score direction is the Crowding Paradox.}
\label{tab:density_auroc}
\vskip 0.1in
\centering
\small
\begin{tabular}{@{}lcccccc@{}}
\toprule
\multirow{2}{*}{Method} & \multicolumn{2}{c}{$H{=}5\,$d} & \multicolumn{2}{c}{$H{=}10\,$d} & \multicolumn{2}{c}{$H{=}20\,$d} \\
& AUROC & AUPRC & AUROC & AUPRC & AUROC & AUPRC \\
\midrule
GMM density          & \textbf{.941} & \textbf{.094} & \textbf{.943} & \textbf{.176} & \textbf{.961} & \textbf{.488} \\
Fragility Score      & .941 & .096 & .942 & .167 & .960 & .460 \\
HMM ($k{=}3$)        & .920 & .061 & .886 & .078 & .770 & .119 \\
\midrule
Isolation Forest     & .390 & .010 & .306 & .013 & .412 & .039 \\
OCSVM                & .115 & .005 & .294 & .013 & .443 & .041 \\
LOF                  & .532 & .010 & .466 & .017 & .527 & .046 \\
PCA reconstruction   & .802 & .023 & .715 & .033 & .606 & .057 \\
AE reconstruction    & .431 & .009 & .364 & .014 & .535 & .061 \\
\bottomrule
\end{tabular}
\vskip -0.1in
\end{table}

GMM density and Fragility Score consistently outperform every anomaly baseline by a wide margin (AUROC $\ge .94$ at all horizons; best anomaly baseline reaches $.80$ at $H{=}5$ and degrades thereafter).
The dominance is even sharper on AUPRC, which is more informative for the rare-event regime.
This empirically validates the central premise of the paper: density, not anomaly, is the operative risk signal for crowded multi-agent systems, and motivates using the density-derived Fragility Score as the QUBO reward.

\section{Risk-Diversity Pareto Front}
\label{app:pareto}

\begin{table}[H]
\caption{Full solver Pareto results at $n{=}15$, $K{=}8$, horizon${=}$5d (extends Table~\ref{tab:main} with all QAOA variants). SA/Exact: 5 data-resampling seeds. QAOA-sim: 5 classical-optimizer seeds. QAOA-Rig (Ankaa-3): single submission per $\lambda$ ($\$0.75$/task; device retired during this work). QAOA-Rig (Cepheus-1-108Q): 3 independent QPU submissions per $\lambda$ ($\$0.51$/task).}
\label{tab:pareto_full}
\vskip 0.1in
\centering
\small
\begin{tabular}{@{}llccc@{}}
\toprule
$\lambda$ & Method & Avg FS & Diversity & $|\mathcal{S}|$ \\
\midrule
\multirow{5}{*}{0.0}
& Greedy & .367\scriptsize$\pm$.002 & .476\scriptsize$\pm$.035 & 8.0 \\
& SA & .366\scriptsize$\pm$.002 & .487\scriptsize$\pm$.029 & 8.0 \\
& Exact & .367\scriptsize$\pm$.002 & .476\scriptsize$\pm$.035 & 8.0 \\
& QAOA-sim & .367\scriptsize$\pm$.000 & .506\scriptsize$\pm$.057 & 8.0 \\
& QAOA-Rig (Ankaa-3) & .368 & .481 & 8 \\
\midrule
\multirow{6}{*}{0.5}
& Greedy & .367\scriptsize$\pm$.002 & .476\scriptsize$\pm$.035 & 8.0 \\
& SA & .348\scriptsize$\pm$.025 & .591\scriptsize$\pm$.070 & 8.0 \\
& Exact & .348\scriptsize$\pm$.025 & .591\scriptsize$\pm$.070 & 8.0 \\
& QAOA-sim & .366\scriptsize$\pm$.001 & .586\scriptsize$\pm$.027 & 8.0 \\
& QAOA-Rig (Ankaa-3) & .364 & .611 & 8 \\
& QAOA-Rig (Cepheus-1) & .366\scriptsize$\pm$.001 & .592\scriptsize$\pm$.020 & 8.0 \\
\midrule
\multirow{6}{*}{1.0}
& Greedy & .367\scriptsize$\pm$.002 & .476\scriptsize$\pm$.035 & 8.0 \\
& SA & .337\scriptsize$\pm$.032 & .639\scriptsize$\pm$.062 & 7.2 \\
& Exact & .337\scriptsize$\pm$.032 & .639\scriptsize$\pm$.062 & 7.2 \\
& QAOA-sim & .365\scriptsize$\pm$.002 & .628\scriptsize$\pm$.016 & 7.0 \\
& QAOA-Rig (Ankaa-3) & .362 & .640 & 7 \\
& QAOA-Rig (Cepheus-1) & .367\scriptsize$\pm$.001 & .616\scriptsize$\pm$.007 & 7.0 \\
\bottomrule
\end{tabular}
\vskip -0.1in
\end{table}

\begin{figure}[H]
\centering
\includegraphics[width=0.55\textwidth]{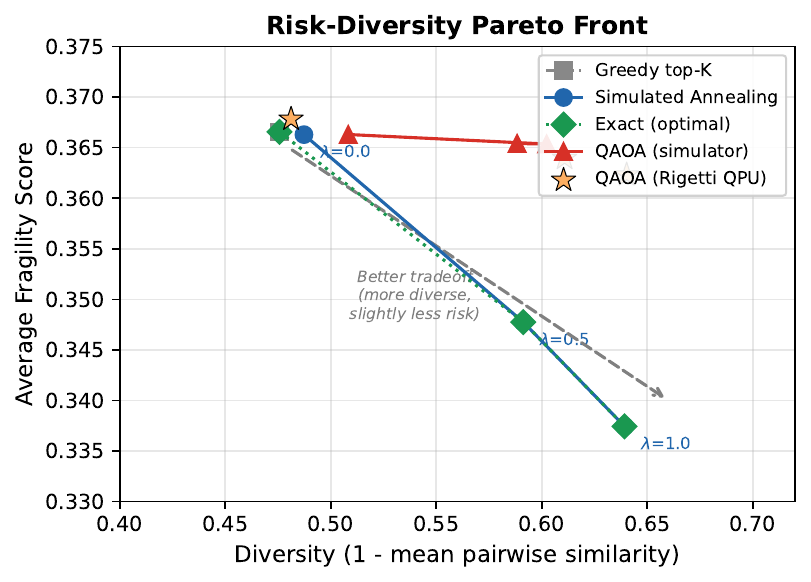}
\caption{Risk-diversity Pareto front across $\lambda \in \{0.0, 0.5, 1.0\}$ at $n{=}15$, $K{=}8$, horizon${=}$5d.
Greedy (gray) is fixed regardless of $\lambda$.
SA and Exact trace the optimal Pareto curve; QAOA on Rigetti (Ankaa-3 single-seed and Cepheus-1-108Q multi-seed) tracks the frontier within hardware noise.}
\label{fig:pareto}
\end{figure}

Greedy selection is Pareto-dominated: QUBO-aware methods achieve higher diversity with competitive Avg FS.
The $\lambda$ parameter provides smooth, interpretable control along the frontier.
QAOA solutions, both simulated and hardware, lie close to the SA/Exact frontier; the multi-seed Cepheus-1-108Q runs (FS $.366\pm.001$ / Div $.592\pm.020$ at $\lambda{=}0.5$, FS $.367\pm.001$ / Div $.616\pm.007$ at $\lambda{=}1.0$) sit slightly inside the Exact frontier as expected from the QAOA $p{=}1$ approximation plus 500-shot measurement noise.

\paragraph{Solver comparison at $\lambda{=}0.5$.}
\begin{figure}[H]
\centering
\includegraphics[width=0.55\textwidth]{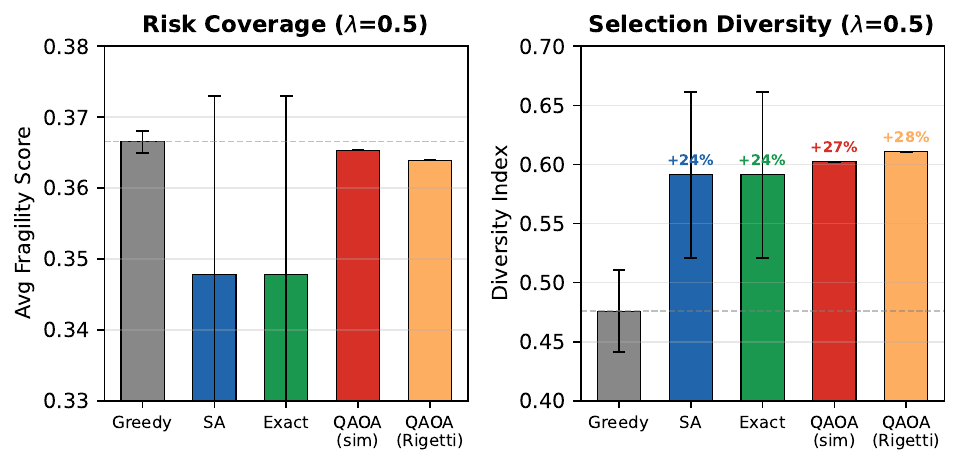}
\caption{Solver comparison at $\lambda{=}0.5$.
Left: Avg Fragility Score. Right: Diversity.
Percentages denote improvement over the Greedy baseline.
Error bars: $\pm 1$ std across 5 seeds.}
\label{fig:bars}
\end{figure}
SA and Exact achieve identical solutions ($+24\%$ diversity over Greedy).
QAOA-simulator and QAOA-Rig at $\lambda{=}0.5$ achieve $+24\%$ to $+28\%$ diversity respectively, while maintaining FS within $1\%$ of the Greedy baseline.

\section{Cross-Horizon Consistency}
\label{app:horizons}

\begin{figure}[H]
\centering
\includegraphics[width=0.55\textwidth]{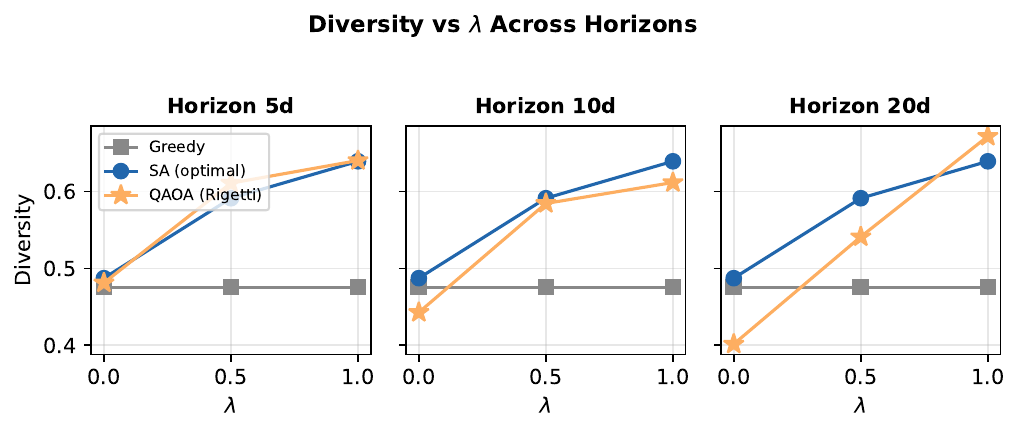}
\caption{Diversity as a function of $\lambda$ across crash horizons (5d, 10d, 20d).
SA consistently increases diversity with $\lambda$; Rigetti results follow the same trend with higher variance.}
\label{fig:horizons}
\end{figure}

The QUBO objective depends on FS and similarity rather than on horizon-specific labels, so the optimization is horizon-agnostic by construction.
The variation in QAOA-Rig results across horizons reflects finite sampling (500 shots) and hardware noise rather than systematic horizon dependence.

\begin{table}[H]
\caption{Cross-horizon results at $\lambda{=}0.5$ for Greedy, SA, and QAOA-Rig (Ankaa-3 single-seed).}
\label{tab:horizons}
\vskip 0.1in
\centering
\small
\begin{tabular}{@{}llccc@{}}
\toprule
Horizon & Method & Avg FS & Diversity & $|\mathcal{S}|$ \\
\midrule
\multirow{3}{*}{5d}
& Greedy & .367 & .476 & 8 \\
& SA/Exact & .348 & .591 & 8 \\
& QAOA-Rig & .364 & .611 & 8 \\
\midrule
\multirow{3}{*}{10d}
& Greedy & .367 & .476 & 8 \\
& SA/Exact & .348 & .591 & 8 \\
& QAOA-Rig & .366 & .584 & 8 \\
\midrule
\multirow{3}{*}{20d}
& Greedy & .367 & .476 & 8 \\
& SA/Exact & .348 & .591 & 8 \\
& QAOA-Rig & .367 & .541 & 8 \\
\bottomrule
\end{tabular}
\end{table}

\section{Scaling Behavior ($n{=}15$ to $30$)}
\label{app:scaling}

\begin{figure}[H]
\centering
\includegraphics[width=0.55\textwidth]{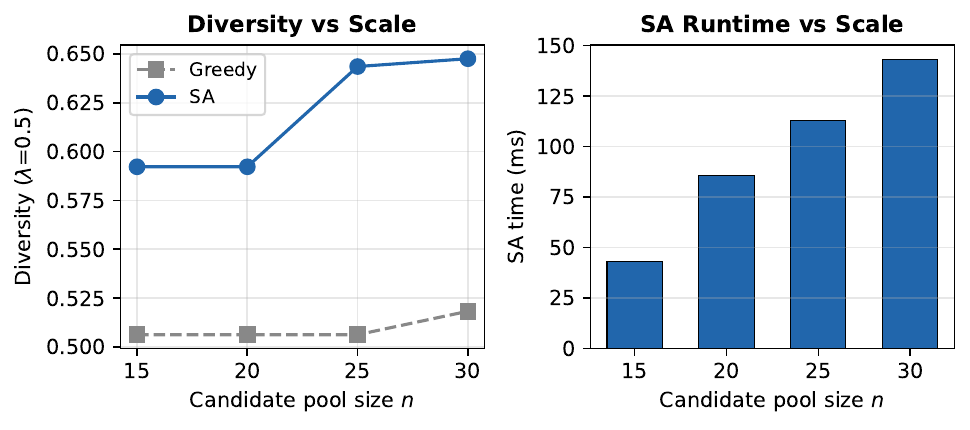}
\caption{Left: SA diversity gain over Greedy remains stable as the candidate pool grows from $n{=}15$ to $30$ ($\lambda{=}0.5$).
Right: SA runtime scales from 43ms to 143ms.}
\label{fig:scaling}
\end{figure}

The diversity advantage of QUBO-based optimization is not an artifact of small $n$.
As the candidate pool grows from 15 to 30, SA consistently achieves diversity $\sim$0.59--0.65 at $\lambda{=}0.5$, compared with Greedy's $\sim$0.51--0.52.
SA runtime remains practical, growing from 43ms ($n{=}15$) to 143ms ($n{=}30$).
We also executed QAOA on Rigetti Ankaa-3 at $n{=}20$ (630-gate circuit, \$0.75/task), obtaining diversity 0.577 at $\lambda{=}0.5$, confirming that the formulation remains executable on real quantum hardware beyond the smallest verifiable scale.

\paragraph{Extended scaling to $n{=}200$.}
To verify that the QUBO advantage is not an artifact of small problem size, we scale the candidate pool to $n \in \{50, 100, 200\}$ on the full financial test set.
At $n{=}200$ with $\lambda{=}0.5$, SA achieves diversity $0.851$ compared with Greedy's $0.514$, a $+66\%$ improvement.
The diversity gain grows monotonically with $n$: $+24\%$ at $n{=}15$, $+31\%$ at $n{=}50$, $+46\%$ at $n{=}100$, and $+66\%$ at $n{=}200$.
SA runtime grows from 44ms ($n{=}15$) to 56 seconds ($n{=}200$), remaining practical for daily monitoring.

\section{Sensitivity to $K$ and $\lambda$}
\label{app:ksens}

\begin{figure}[H]
\centering
\includegraphics[width=0.55\textwidth]{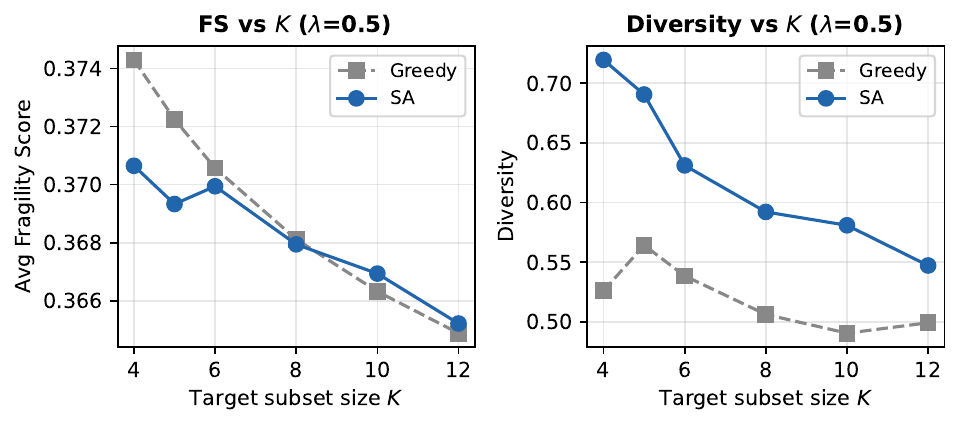}
\caption{Diversity and FS as a function of target subset size $K$ ($n{=}15$, $\lambda{=}0.5$).
SA outperforms Greedy in diversity across all $K$ values.}
\label{fig:ksweep}
\end{figure}

\begin{figure}[H]
\centering
\includegraphics[width=0.55\textwidth]{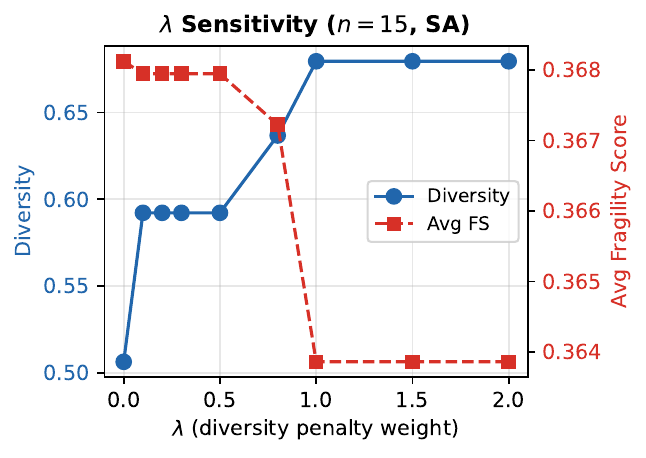}
\caption{Fine-grained $\lambda$ sweep ($n{=}15$, SA).
Diversity (blue) exhibits a sharp onset near $\lambda{\approx}0.1$ and saturates beyond $\lambda{\approx}1.0$.
FS (red) decreases monotonically but remains within $1\%$ of baseline for $\lambda{\le}0.5$.}
\label{fig:lambdasweep}
\end{figure}

The SA-over-Greedy diversity advantage holds across target subset sizes $K \in \{4, 5, 6, 8, 10, 12\}$, confirming that the result is not specific to $K{=}8$.
The gain is most pronounced at smaller $K$, where selecting few diverse states is combinatorially hardest.
The $\lambda$ sweep shows a sharp diversity onset near $\lambda{\approx}0.1$ -- where the similarity penalty first becomes strong enough to alter the selection -- followed by saturation beyond $\lambda{\approx}1.0$.
FS decreases monotonically with $\lambda$ but remains within $1\%$ of the Greedy baseline for $\lambda{\le}0.5$, indicating a favorable operating region.
Pilot experiments with cosine similarity yield qualitatively similar behavior, suggesting that the effect is not kernel-specific.

\section{Learned $\lambda$ Policy}
\label{app:learn}

A fixed $\lambda$ ignores the structure of each candidate pool.
We train a lightweight MLP policy ($10$-dim pool statistics $\to \lambda$) via REINFORCE, optimizing a composite reward of FS, diversity, and cardinality satisfaction.

\begin{table}[H]
\caption{Learned $\lambda$ on held-out test episodes. Objective is the composite reward used during training.}
\label{tab:learn}
\vskip 0.1in
\centering
\small
\begin{tabular}{@{}lcccc@{}}
\toprule
Method & Objective & FS & Diversity & $\hat\lambda$ \\
\midrule
Greedy & 0.908 & .406 & .502 & 0 \\
Fixed $\lambda{=}0.5$ & 1.055 & .396 & .660 & 0.50 \\
Learned $\lambda$ & \textbf{1.071} & .394 & .677 & 0.64 \\
Grid (oracle) & 1.076 & .393 & .684 & 0.78 \\
\bottomrule
\end{tabular}
\end{table}

The learned policy achieves $99.5\%$ of the oracle grid-search objective on held-out test data, generalizing from the training period (pre-2022) to the test period (post-2024) without retraining.

\section{Cross-Domain Generalization}
\label{app:cross}

\begin{table}[H]
\caption{Cross-domain QUBO results at $\lambda{=}0.5$, $K{=}8$, $n{=}50$, 3 seeds. Diversity is $1-\bar{S}$ on the domain-specific RBF kernel; absolute values are not comparable across domains because the kernel bandwidth differs.}
\label{tab:cross}
\vskip 0.1in
\centering
\small
\begin{tabular}{@{}lccc@{}}
\toprule
Domain & Greedy Div & SA Div & Relative gain \\
\midrule
Financial      & .518 & .657\scriptsize$\pm$.028 & $+$27\% \\
Traffic (inD)$^\dagger$  & .000 & .421\scriptsize$\pm$.012 & saturated \\
MARL (SMAC v2) & .763 & .816\scriptsize$\pm$.009 & $+$7\% \\
\bottomrule
\end{tabular}
\vskip 0.05in
\footnotesize
$^\dagger$Traffic Greedy diversity rounds to $.000$ ($.0001$ before rounding) because the top-FS candidate pool consists of consecutive frames at the same crowded intersection, which RBF-saturate to similarity $\approx 1$ at the chosen bandwidth; SA spreads selections across distinct configurations and recovers measurable diversity.
\end{table}

The QUBO formulation provides positive diversity gains over Greedy in all three domains, but the magnitude varies: large in finance and traffic, modest in MARL where Greedy already selects relatively diverse episodes ($.763$).
The Traffic Greedy diversity rounding to zero is a genuine RBF saturation effect, not a measurement bug, and is itself the strongest case for moving from score-and-threshold to QUBO subset selection: Greedy returns a degenerate redundant subset, while SA returns a useful one.

\section{Downstream Stress Capture}
\label{app:downstream}

To break the tautological coupling between the QUBO objective (similarity penalty) and the in-paper diversity metric (1 minus mean similarity), we evaluate each method on \emph{forward-looking} market metrics that are independent of the optimization signal: realized 5-day forward drawdown of SPY at the selected dates, realized 5-day forward volatility, temporal coverage (number of selections that are at least 7 trading days apart), and recall on the labeled \texttt{structural\_break} events at a 20-day window.

\begin{table}[H]
\caption{Downstream stress-capture metrics on the financial test set ($n{=}15$, $K{=}8$, 5 seeds; mean$\pm$std). Forward drawdown is the realized 5-day max drawdown of SPY at the selected dates (more negative = stress days successfully captured); recall is computed against the 73 labeled structural-break events.}
\label{tab:downstream}
\vskip 0.1in
\centering
\small
\begin{tabular}{@{}lcccc@{}}
\toprule
Method & Fwd DD-5d (\%) & Fwd vol-5d (\%) & Cov.\ 7d & Recall (SB,20d) \\
\midrule
Greedy            & $-0.90$\scriptsize$\pm$.13 & $0.50$\scriptsize$\pm$.06 & 5.4 & .05 \\
SA, $\lambda{=}0.5$    & $-0.87$\scriptsize$\pm$.06 & $0.56$\scriptsize$\pm$.04 & 5.6 & \textbf{.13} \\
SA, $\lambda{=}1.0$    & $-0.70$\scriptsize$\pm$.19 & $0.54$\scriptsize$\pm$.06 & 5.4 & .10 \\
MMR$(0.5)$        & $-1.17$\scriptsize$\pm$.17 & $0.65$\scriptsize$\pm$.05 & 4.8 & .10 \\
MMR$(0.7)$        & $\mathbf{-1.22}$\scriptsize$\pm$.22 & $0.64$\scriptsize$\pm$.06 & 5.2 & \textbf{.15} \\
$k$-DPP           & $-1.01$\scriptsize$\pm$.27 & $0.61$\scriptsize$\pm$.05 & 5.0 & .08 \\
\bottomrule
\end{tabular}
\end{table}

The downstream metrics support the framing of QUBO as a controllable frontier rather than as a single operating point.
At $\lambda{=}0.5$, QUBO-SA more than doubles the structural-break recall over Greedy ($.13$ versus $.05$) while preserving comparable temporal coverage, and at higher $\lambda$ the same formulation reaches a realized forward drawdown of $-0.70\%$.
MMR at diversity weight $0.7$ achieves $-1.22\%$ realized drawdown, which is a strong outcome but is obtained at a single fixed operating point, because MMR exposes only its diversity weight $\mu$ and reaching a different stress regime would require switching to another method.
The QUBO formulation, in contrast, dials the same property through $\lambda$ along a continuous Pareto curve that contains MMR and $k$-DPP as fixed points; once a deployment specifies which downstream metric matters (recall, drawdown, coverage, or a weighted combination), the appropriate $\lambda$ is read off the frontier rather than chosen between methods.
The claim our experiments therefore support is one of generality across operating points, rather than of point-by-point dominance over any individual baseline.

\section{Solver Computational Cost}
\label{app:cost}

\begin{table}[H]
\caption{Solver cost comparison per QUBO instance ($n{=}15$).}
\label{tab:cost}
\vskip 0.1in
\centering
\small
\begin{tabular}{@{}lccc@{}}
\toprule
Solver & Time & Hardware & USD Cost \\
\midrule
Greedy & $<$1ms & CPU & \$0 \\
SA (200 reads) & 45ms & CPU & \$0 \\
Exact & 26ms & CPU & \$0 \\
QAOA-sim & $\sim$7s & CPU & \$0 \\
QAOA-Rig (Ankaa-3) & $\sim$2min$^*$ & QPU & \$0.75 \\
QAOA-Rig (Cepheus-1) & $\sim$40s$^*$ & QPU & \$0.51 \\
\bottomrule
\multicolumn{4}{l}{\scriptsize $^*$Includes local param.\ optimization $+$ QPU queue $+$ 500 shots.}
\end{tabular}
\end{table}

\section{Why QUBO over Gradient Methods, and QAOA as Forward-Looking Backend}
\label{app:analysis}

\paragraph{Why QUBO over gradient methods?}
Subset selection is a discrete problem: the decision variables $z_i \in \{0,1\}$ do not admit continuous relaxation without introducing approximation.
Continuous relaxation methods such as Gumbel-Softmax or straight-through estimators struggle to enforce exact cardinality and pairwise diversity constraints simultaneously, and typically require heuristic rounding that breaks optimality guarantees.
QUBO encodes the entire objective, including the cardinality constraint via the penalty term $({\sum_i z_i - K})^2$, as a single polynomial~\citep{glover2022tutorial}.
The same instance can be dispatched to any compatible solver without reformulation, from classical SA to quantum annealers to gate-model QAOA.

\paragraph{QAOA as a forward-looking backend.}
The goal of this work is not to demonstrate quantum advantage at small scale but to verify that the same decision formulation executes on real quantum hardware without modification.
At $n{=}15$, classical SA finds the exact optimum in 45ms on a single CPU core; quantum advantage is not expected at this scale.
We verify \emph{computational compatibility}: the QUBO formulation provides a natural interface to quantum hardware via the standard QAOA ansatz, requiring no problem-specific circuit design.
The multi-seed Cepheus-1-108Q evaluation (Sec.~\ref{sec:experiments}) shows the per-seed std at $\lambda{=}1.0$ is $.007$, while the gap between the multi-seed mean ($.616$) and Exact ($.639$) is $.023$, suggesting the residual gap is dominated by the $p{=}1$ ansatz approximation rather than measurement noise; raising $p$ on simulator and on hardware (subject to coherence-time budgets) is the natural next experiment.
We further extend QAOA execution to $n{=}20$ on Rigetti Ankaa-3 (630-gate circuit), confirming the formulation remains executable beyond the smallest verifiable scale.
As quantum hardware scales, larger QUBO instances from denser MAS monitoring scenarios could benefit from quantum parallelism, while the problem formulation and evaluation framework remain unchanged.

\end{document}